\documentclass[sigconf]{acmart}

\AtBeginDocument{%
  \providecommand\BibTeX{{%
    \normalfont B\kern-0.5em{\scshape i\kern-0.25em b}\kern-0.8em\TeX}}}

\copyrightyear{2024}
\acmYear{2024}
\setcopyright{acmlicensed}\acmConference[SIGGRAPH Conference Papers '24]{Special Interest Group on Computer Graphics and Interactive Techniques Conference Conference Papers '24}{July 27-August 1, 2024}{Denver, CO, USA}
\acmBooktitle{Special Interest Group on Computer Graphics and Interactive Techniques Conference Conference Papers '24 (SIGGRAPH Conference Papers '24), July 27-August 1, 2024, Denver, CO, USA}
\acmDOI{10.1145/3641519.3657489}
\acmISBN{979-8-4007-0525-0/24/07}

\usepackage{color}
\usepackage{bm}
\usepackage{soul}
\usepackage{placeins}
\usepackage{dblfloatfix}

\usepackage{algorithm}
\usepackage{algpseudocode}

\usepackage{caption}
\usepackage{subcaption}
\definecolor{mathbrace_color}{rgb}{0.2,0.5,1.0}

\newcommand{\beq}{\begin{equation}}
\newcommand{\eeq}{\end{equation}}

\newcommand{\bracetext}[1]{\text{\footnotesize #1}}

\newcommand{\cunderbrace}[2]{\color{mathbrace_color}\underbrace{\color{black}#1}_{\bracetext{#2}}\color{black}}
\newcommand{\cunderbracelines}[3]{\color{mathbrace_color}\underbrace{\color{black}#1}_{\substack{\bracetext{#2}\\\bracetext{#3}}}\color{black}}

\newcommand{\coverbracelines}[3]{\color{mathbrace_color}\overbrace{\color{black}#1}^{\substack{\text{\scriptsize#2}\\\text{\scriptsize#3}}}\color{black}}

\newcommand{\coverlabel}[2]{\color{white}\overbrace{\color{black}#1}^{\bracetext{\color{mathbrace_color}#2}}\color{black}}

\newcommand{\coverlabellines}[3]
{\color{white}\overbrace{\color{black}#1}^{\substack{\text{\scriptsize{\color{mathbrace_color}#2}}\\\text{\scriptsize{\color{mathbrace_color}#3}}}}\color{black}}

\newcommand{\namedparagraph}[1]{\vspace{0.1cm}\noindent\textbf{#1}}

\newcommand{\mb}[1]{\mathbf{#1}}
\newcommand{\vect}[1]{\vec{\mathbf{#1}}}
\newcommand{\sample}{x}
\newcommand{\fsample}[1]{\sample_{#1}} %
\newcommand{\rsample}[1]{\hat{\sample{}}_{#1}} %
\newcommand{\xsignal}[1]{s_{#1}}
\newcommand{\restimate}[1]{\xsignal{#1}}
\newcommand{\rpreguideestimate}[1]{\xsignal{#1}'}

\newcommand{\gsample}[1]{\hat{\sample{}}_{#1}} %
\newcommand{\datadist}{q} %

\newcommand{\anyimage}{\mb{x}}

\newcommand{\gpreguidesample}[1]{\gsample{#1}'}

\newcommand{\dstructv}{\vect{d}_g}
\newcommand{\dstruct}[1]{\vect{d}_g(#1)}

\newcommand{\adjustmentweight}{\delta}

\newcommand{\rvar}[1]{\rstd{#1}^2}
\newcommand{\rstd}[1]{\sigma_{#1}}
\newcommand{\stdnrm}{\mathbf{z}}
\newcommand{\rmean}{\mathbf{\mu}_{\theta}}

\newcommand{\guideim}{\fsample{g}}

\newcommand{\guideprop}{\mathbf{f}}
\newcommand{\guidepropof}[1]{\mathbf{f}(#1)}

\newcommand{\normdist}[3]{\mathcal{N}(#2,#3)}
\newcommand{\stdnormdist}{\mathcal{N}(0,1)}

\newcommand{\dmodel}{\mathbf{\epsilon}_\theta}

\newcommand{\fbeta}[1]{\beta_{#1}} %
\newcommand{\falpha}[1]{\alpha_{#1}} %
\newcommand{\fnoise}[1]{\mathbf{z}_{#1}} %
\newcommand{\falphacum}[1]{\tilde{\alpha}_{#1}}

\newcommand{\ITIT}{I2I}

\newcommand{\tstop}{t_{end}}

\begin{document}

\title{Filter-Guided Diffusion for Controllable Image Generation}

\author{Zeqi Gu}
\authornote{Both authors contributed equally to this research.}
\affiliation{%
  \institution{Cornell Tech}
  \streetaddress{2 W Loop Rd}
  \city{New York, NY}
  \country{USA}}
\email{zg45@cornell.edu}

\author{Ethan Yang}
\authornotemark[1]
\affiliation{%
  \institution{Cornell University}
  \streetaddress{107 Hoy Rd}
  \city{Ithaca, NY}
  \country{USA}}
\email{eey8@cornell.edu}

\author{Abe Davis}
\affiliation{%
  \institution{Cornell University}
  \streetaddress{107 Hoy Rd}
  \city{Ithaca, NY}
  \country{USA}}
\email{abedavis@cornell.edu}

\renewcommand{\shortauthors}{Gu and Yang, et al.}

\begin{abstract}
Recent advances in diffusion-based generative models have shown incredible promise for zero shot image-to-image translation and editing. Most of these approaches work by combining or replacing network-specific features used in the generation of new images with those taken from the inversion of some guide image. Methods of this type are considered the current state-of-the-art in training-free approaches, but have some notable limitations: they tend to be costly in runtime and memory, and often depend on deterministic sampling that limits variation in generated results. We propose Filter-Guided Diffusion (FGD), an alternative approach that leverages fast filtering operations during the diffusion process to support finer control over the strength and frequencies of guidance and can work with non-deterministic samplers to produce greater variety. With its efficiency, FGD can be sampled over multiple seeds and hyperparameters in less time than a single run of other SOTA methods to produce superior results based on structural and semantic metrics. We conduct extensive quantitative and qualitative experiments to evaluate the performance of FGD in translation tasks and also demonstrate its potential in localized editing when used with masks.
\end{abstract}



\begin{teaserfigure}
    \centering
    \captionsetup{type=figure}
    \includegraphics[width=\textwidth]{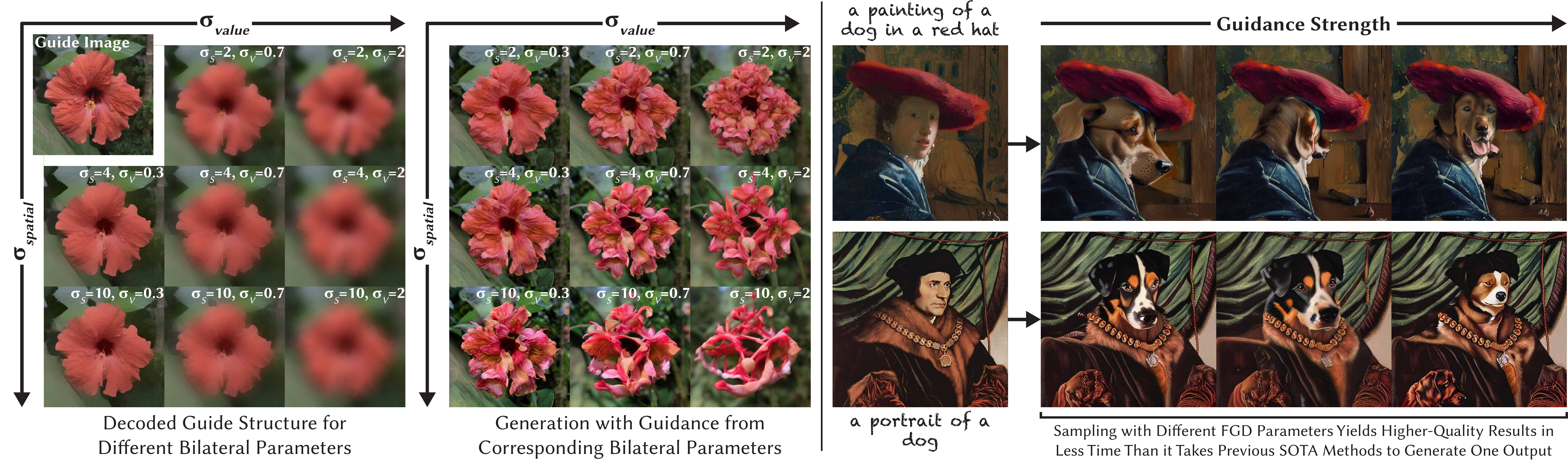}
    \captionof{figure}{
    We present Filter-Guided Diffusion (FGD), a training-free black box method for conditioning image diffusion on the structure of example images. FGD works by adding a fast filtering step between each iteration of the diffusion process. Building on classic image processing theory, we can design this filtering step to preserve the structure of a given guide image. The grid on the far left shows the effects of classic bilateral filters with different parameters on an image. The middle grid shows the result of using the same filters for guidance with FGD (in this case, using an empty prompt). The spatial filter size, which controls the scale of blur in a traditional bilateral, determines the spatial scale at which the network may vary from structure of the input image. The value filter size, which controls edge preservation in the traditional bilateral, determines how much the diffusion process should respect the edges of the guide. In addition to the filter's parameters, we can control the overall strength of guidance to determine how closely the result should resemble the input. The right shows two examples of image-to-image translation with varying guidance strength. Results become closer to the guide image as guidance strength increases to the right.}
    \label{fig:teaser}
\end{teaserfigure}%

\maketitle

\begingroup
\renewcommand\thefootnote{}
\footnotetext{\copyright~2024 Association for Computing Machinery.  
This is the author's version of the work. It is posted here for your personal use.  
Not for redistribution.  
The definitive Version of Record was published in \textit{ACM SIGGRAPH Conference Papers '24},  
\url{https://doi.org/10.1145/3641519.3657489}}
\addtocounter{footnote}{0}
\endgroup

\section{Introduction}
\label{sec:intro}
In recent years, text-conditioned diffusion models have become the basis for state-of-the-art (SOTA) solutions to a range of image generation tasks. While these models natively synthesize images based on textual descriptions, several works have explored ways to adapt them to Image-to-Image translation (\ITIT{}) tasks. While early efforts on zero-shot I2I with diffusion models explored \emph{black box} approaches---meaning methods that are architecture-independent and do not require access to the internal feature of a network---more recent SOTA zero-shot approaches have leveraged attention features derived from a costly inversion process based on Denoising Diffusion Implicit Models (DDIM)~\cite{song2020denoising} with Stable Diffusion (SD)~\cite{rombach2022high}.
In this paper, we show that much of the quality gained by these architecture-dependent
approaches can be achieved with a much faster filtering strategy, which we call \emph{Filter-Guided Diffusion} (FGD).
Our approach is based on the observation that we can make a diffusion process selectively preserve certain information by filtering the denoised images estimated at each diffusion step. Building on older image processing techniques like bilateral filtering and Laplacian blending, we derive a fast linearization of the joint bilateral filter to preserve the structure of example images in generated outputs.
We show that the speed and diversity achieved by FGD allow it to produce higher-quality results in less time than existing zero-shot methods by simply sampling from multiple seeds or filter parameters at runtime.

\section{Related Work}
\label{sec:relatedwork}
\subsection{Early Image-to-Image Translation Methods}

Early methods for \ITIT{} used linear filters to selectively blend or transfer information from specific frequency bands in an image (e.g., ~\cite{LaplacianPyramid,hybridimages}). This was later improved with the use of edge-aware filters like the bilateral~\cite{bilateraltomasi} to perform photographic style transfer~\cite{BilateralStyleTransfer06}. 
Following these early works, the development of Generative Adversarial Networks (GANs), led to approaches that learn a mapping between image distributions directly from training data ~\cite{pix2pix2017,CycleGAN2017,park2020contrastive,karras2019style,wei2021simple,  karras2017progressive, chen2009sketch2photo, parmar2022spatially}. More recently, methods based on large pre-trained diffusion models have been shown to outperform GANs on most \ITIT{} tasks. We divide our discussion of these methods into two categories: \emph{black box} methods, which are zero-shot approaches that do not depend on internal features of any specific network architecture, as opposed to \emph{white box} methods, which typically require additional fine-tuning or architecture-specific features derived from an inversion process.

\subsection{Black Box Diffusion Approaches}
Two of the earliest black box \ITIT{} diffusion approaches were SDEdit~\cite{meng2021sdedit} and ILVR~\cite{choi2021ilvr}. SDEdit works by simply running the latter portion of the reverse diffusion process on a partially noised guide image. ILVR conditions the generation of new images on a downsampled version of the guide by replacing low frequencies from the output of each diffusion step with the corresponding low frequencies of the guide. FGD operates on a similar principle, but improves upon ILVR in several significant ways. Notably, we operate on the space of denoised images instead of the noisy intermediate outputs, and our filtering operation is designed to preserve edge structure while also allowing the distribution of colors to adapt with given prompts. We highlight these differences in Sec.~\ref{sec:distinctions_from_others}

\subsection{White Box Diffusion Approaches}
Most recent works approach \ITIT{} diffusion using white box methods. 
Several do this by fine-tuning diffusion, either for the target domain specified by the prompt ~\cite{kim2022diffusionclip}, or to preserve attributes of the guide image~\cite{kawar2023imagic, kwon2022diffusion}. Another popular class of approaches leverages network-specific attention features generated through a DDIM inversion~\cite{song2020denoising} of the guide image. This results in inheriting guide structure, but limits generation diversity due to DDIM being a deterministic process. For example, Plug-and-Play (PnP)~\cite{tumanyan2022plug} preserves structure by substituting self-attention features produced through inversion.
Prompt-to-Prompt (P2P)~\cite{hertz2022prompt} and Pix2Pix-Zero (P2P0)~\cite{parmar2023zero} also use DDIM inversion to extract attention features, and require additional source text describing the guide image to drive edits. P2P uses source and target text to locate regions for editing and manipulates the attention masks for these regions according to specific types of translation tasks. P2P0 uses source and target text to calculate a CLIP direction for edits, then optimizes for adherence to cross-attention maps derived from inversion during generation. Closely related to these lines of work is InstructPix2Pix~\cite{brooks2023instructpix2pix} which performs \ITIT{} based on editing instructions instead of a common target prompt. They accomplishes this by fine-tuning a conditional diffusion model on over 450,000 synthetic image-text instruction pairs generated with P2P and GPT-3~\cite{brown2020language}. In contrast to these white box methods, our method requires no source prompts, no fine-tuning or retraining, and no need for inversion or knowledge of model-specific attention features, which saves considerable time and memory.

There are also methods that focus on improvements orthogonal to ours, such as optimizing starting noise for editing~\cite{mokady2023null}. Refer to our supplemental for a more complete discussion.

\section{Diffusion Background}
\label{sec:background}
We begin with a brief review of the general diffusion process. Note that our derivation differs slightly from others by introducing $\restimate{t}$ to denote the intermediate estimated denoised image at step $t$, which simplifies notation for our method.

\namedparagraph{Forward Process:}
The forward diffusion process takes a sample $\fsample{0}\sim\datadist(\fsample{0})$ from our data distribution and iteratively mixes it with Gaussian noise. We can describe this process in terms of consecutive forward steps $\fsample{t-1}$ and  $\fsample{t}$:
\beq
\datadist(\fsample{t}|\fsample{t-1}):=\normdist{\fsample{t}}{\sqrt{\falpha{t}}\fsample{t-1}}{\fbeta{t}}
\label{eq:diffusionstepclassic}
\eeq
\noindent where 
$\fbeta{t} \in (0,1)$ and $\falpha{t}=1-\fbeta{t}$ provide a schedule of noise added at each step. Unrolling this equation gives:
\beq
\datadist(\fsample{t}|\fsample{0}):=\normdist{\fsample{t}}{\sqrt{\falphacum{t}}\fsample{0}}{1-\falphacum{t}}
\label{eq:forwarddiffusiondirect}
\eeq
\noindent where $\falphacum{t}=\prod_{i=1}^t \falpha{i}$. We can interpret Eq.~\ref{eq:forwarddiffusiondirect} as a linear combination of the initial data sample $\fsample{0}$ and noise $\fnoise{}\sim\stdnormdist{}$ drawn from a standard normal distribution:
\beq
\fsample{t}=\cunderbracelines{(\sqrt{\falphacum{t}})}{signal}{strength}\coverlabel{\fsample{0}}{signal}+\cunderbrace{(\sqrt{1-\falphacum{t}})}{noise strength}\coverlabel{\fnoise{}}{noise}
\label{eq:diffusionstepsignal}
\eeq
Eq.~\ref{eq:diffusionstepsignal} lets us interpret $\fsample{t}$ as a noisy measurement of a signal $\fsample{0}$ with strength $\sqrt{\falphacum{t}}$ and noise strength $\sqrt{1-\falphacum{t}}$, helping us interpret the reverse process in terms of denoising.

\namedparagraph{Reverse Process:} The reverse process is meant to approximate repeated sampling from the posterior distribution $\datadist{}(\fsample{t-1}|\fsample{t})$, which we approximate as Gaussian:
\beq
\datadist(\rsample{t-1}|\rsample{t})\approx\normdist{\rsample{t-1}}{\rmean(\rsample{t})}{\rvar{t}}
\label{eq:reversediffusionstep}
\eeq
To perform this sampling, we need to estimate the means $\rmean(\rsample{t})$ (and optionally, variances $\rvar{t}$) associated with each reverse step.
These estimates come from a network $\dmodel{}$, which is usually trained to estimate the noise $\dmodel{}(\rsample{t})$ associated with a given $\rsample{t}$ (i.e., estimate $\stdnrm{}$ from Eq.~\ref{eq:diffusionstepsignal}). 
This also yields a denoised prediction $\restimate{t}$ of the input signal $\rsample{t}$ at each step. Writing out the connection between our network's input $\rsample{t}$, its output $\dmodel{}(\rsample{t})$, and the intermediate denoised predictions $\restimate{t}$ helps connect our reverse process with the forward one in Eq.~\ref{eq:diffusionstepsignal}:
\beq
\coverlabel{\rsample{t}}{observation}=\cunderbracelines{(\sqrt{\falphacum{t}})}{signal}{strength}
\coverlabellines{\restimate{t}}{signal}{estimate}+\cunderbrace{(\sqrt{1-\falphacum{t}})}{noise strength}
\coverlabellines{\dmodel{}(\rsample{t})}{estimated}{noise}
\label{eq:rdiffusionstepsignal}
\eeq

This connection informs how $\dmodel$ is usually trained: by applying Eq.~\ref{eq:diffusionstepsignal} to training data and predicting $\fnoise{}$. In practice, an iterative approach to estimating $\rsample{0}$ performs better, where the network is used to approximate the means $\rmean(\rsample{t})$ of Eq.~\ref{eq:reversediffusionstep}  at each step: 
\beq
\rmean{}(\rsample{t})=\frac{1}{\sqrt{\falpha{t}}}\left(\rsample{t}-\frac{\fbeta{t}}{\sqrt{1-\falphacum{t}}}\dmodel{}(\rsample{t})\right)
\label{eq:means}
\eeq
\noindent We refer to~\cite{ho2020denoising} for the derivation of Eq.~\ref{eq:means}, which is combined with Eq.~\ref{eq:reversediffusionstep} to compute $\rsample{t-1}$ at each step.

\section{Method}
\label{sec:method}
Our challenge is to balance the goal of guidance against the learned priors of our network. 
We begin by quantifying our guidance objective and considering a trivial solution of that objective. We then proceed with incremental improvements to this trivial solution, highlighting connections with prior work along the way. 

\subsection{Quantifying Guidance}
We define guidance in terms of an input guide image $\guideim$ and filter $\guideprop$. At a high level, the guide filter $\guideprop$ determines what properties we want to guide, and the filtered guide image $\guidepropof{\guideim}$ tells us what values those properties should take. More specifically, we formulate guidance as control over the final generated image $\gsample{0}$ to satisfy:
\beq
\guidepropof{\gsample{0}}\approx\guidepropof{\guideim}
\label{eq:guidanceim}
\eeq

A trivial solution would be to set $\gsample{0}=\guideim{}$. However, this leaves no room for variation that can be used to satisfy other goals (e.g., adherence to a text prompt). To balance guidance against the decisions of the diffusion model, we follow one basic principle: to intervene in the diffusion process {as little and gradually as possible to satisfy Eq.~\ref{eq:guidanceim}. If we assume some smoothness in the behavior of our network, this principle should help ensure that we keep our guided process as close to the network's original training distribution as possible. 
\begin{figure}[!b]
\begin{center}
    \includegraphics[width=1\linewidth]{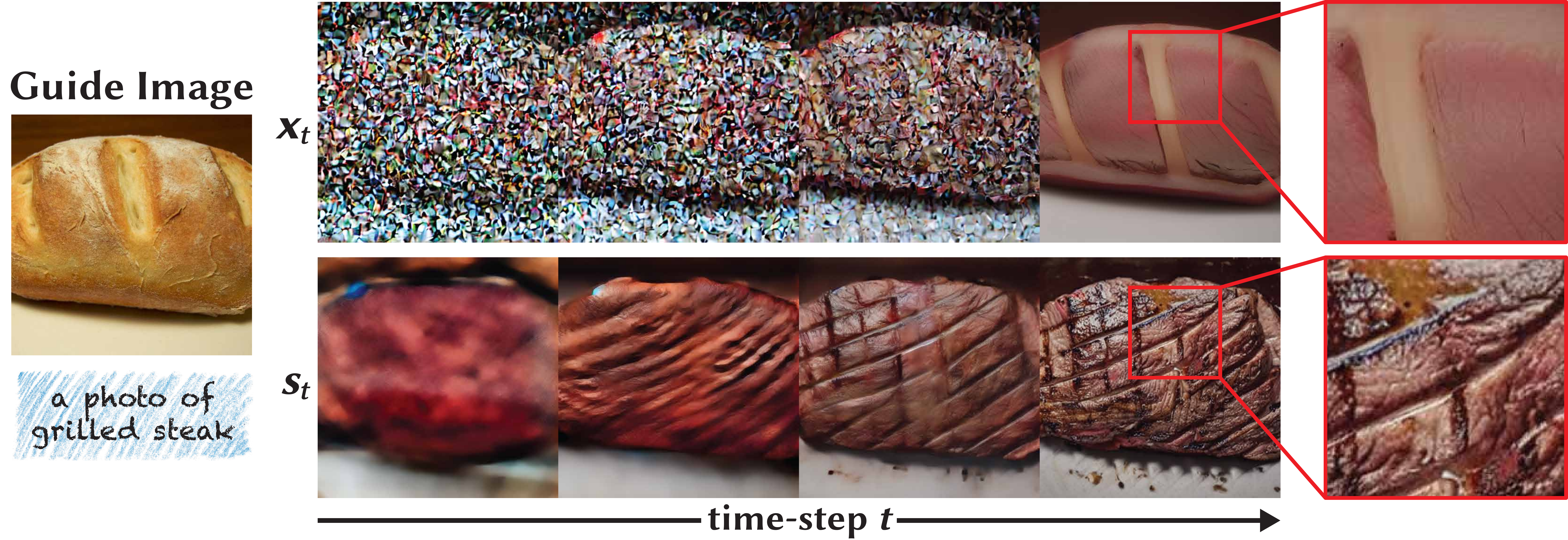}
\end{center}
    \caption{\textbf{Guidance in $s_t$ vs. $x_t$.} Here, we show guidance in $x_t$ (top row) vs. $s_t$ (bottom row), along with the corresponding decoded intermediate latents across time-steps. Since our guide image is closer in distribution to the predicted image $s_t$ than $x_t$, performing filtering in $s_t$ is much less likely to push our diffusion process away from the variance schedule used during training. As a result, observe that guidance in $x_t$ produces over-smoothing artifacts while guidance in $s_t$ is able to add greater detail such as the grill marks while also preserving the structure (last column and inlet).}
\label{fig:stvsxt}
\end{figure}

\subsection{Gradual Guidance}
\begin{figure*}
\begin{center}
    \includegraphics[width=0.85\linewidth]{figures/filter_strength.pdf}
\end{center}
    \caption{\textbf{Effects of $\adjustmentweight$.} Increasing the guidance strength parameter $\adjustmentweight$ causes generated images to take on more of the guide image structure. The results here use a joint bilateral tensor (bottom left) derived from a cat guide image (top left) with $\sigma_{s}=2$, $\sigma_{v}=1$, and $\tstop=15$.
    }
\label{fig:detail_ablation}
\end{figure*}

\label{sec:gradualfiltering}
We proceed by building on our trivial solution $\gsample{0}=\guideim{}$ under the assumption that $\guideprop$ is linear with respect to our image.
In this case, variation of $\rsample{0}$ that lies in the null space of $\guideprop$ will have no effect on Eq.~\ref{eq:guidanceim}. As such, a slightly less trivial solution replaces only the part of $\rsample{0}$ that is orthogonal to this null space:

\beq
\gsample{0}=\rsample{0}-\guidepropof{\rsample{0}}+\guidepropof{\guideim}
\label{eq:laplacianblending}
\eeq
Here we note that, given appropriate choices of $\guideprop$, Eq.~\ref{eq:laplacianblending} can be used to represent older non-neural methods such as Laplacian blending~\shortcite{LaplacianPyramid} or Hybrid Images~\shortcite{hybridimages}.

The main downside of Eq.~\ref{eq:laplacianblending} is that it offers no opportunity for the diffusion process to adapt to our guidance. We can provide this opportunity by applying guidance more gradually at each step, instead of only once at the very end. To simplify notation, we first define the guidance vector $\dstruct{\anyimage{}}$ associated with image vector $\anyimage{}$ as the difference:
\beq
\dstruct{\anyimage{}}=\guidepropof{\guideim}-\guidepropof{\anyimage{}}
\label{eq:guidancevector}
\eeq
\noindent which lets us rewrite Eq.~\ref{eq:laplacianblending} as 
\beq
\gsample{0}=\rsample{0}+\dstruct{\rsample{0}}
\label{eq:filteringst}
\eeq

We can think of $\dstructv$ as a vector that points in the direction of images that satisfy Eq.~\ref{eq:guidanceim}, noting that $\dstruct{\anyimage{}}=0$ whenever this condition is met. If $\gpreguidesample{t-1}$ denotes the unguided output of step $t$, our more gradual strategy will add some multiple $\adjustmentweight{}$ of $\dstructv$ at each step:
\beq
\gsample{t-1}=\cunderbrace{\gpreguidesample{t-1}}{unguided}+\cunderbracelines{\adjustmentweight}{guidance}{strength}\coverbracelines{\dstruct{\gpreguidesample{t-1}}}{guidance}{vector}
\label{eq:guidedupdate}
\eeq

\noindent where $\adjustmentweight$ gives us continuous control over the strength of guidance. Fig.~\ref{fig:detail_ablation} shows the effect of $\adjustmentweight$ on the structure of generated images.

Our gradual guidance strategy has a slight issue; replacing information from a noisy intermediate value $\gsample{t-1}$ with corresponding information taken from a (denoised) guide image $\guideim$ tends to reduce noise overall, which pushes our diffusion process away from the variance schedule used during training.
To avoid this distribution shift, we instead apply our guidance to the estimated denoised image calculated at each step. Following our notation for $\gpreguidesample{}$, we use $\rpreguideestimate{}$ to denote denoised estimates before guidance, and $\restimate{}$ after guidance. From Eq.~\ref{eq:rdiffusionstepsignal} we have
\beq
\rpreguideestimate{t}=\frac{1}{\sqrt{\falphacum{t}}}\left(\rsample{t}-\sqrt{1-\falphacum{t}}\;\dmodel{}(\rsample{t})\right)
\label{eq:st}
\eeq
and from Eq.~\ref{eq:filteringst} we get
\beq
\restimate{t}=\restimate{t}+\adjustmentweight\;\dstruct{\restimate{t}}
\label{eq:guidedupdate_st}
\eeq

Finally, using Eq.~\ref{eq:reversediffusionstep}, 
we can then re-express $\gpreguidesample{t-1}$ in terms of $\restimate{t}$ and $\rsample{t}$ as:
\beq
\gpreguidesample{t-1}=\frac{\fbeta{t}\sqrt{\falphacum{t-1}}}{1-\falphacum{t}}\restimate{t}+\frac{(1-\falphacum{t-1})\sqrt{\falpha{t}}}{1-\falphacum{t}}\rsample{t}+\rvar{t}\stdnrm{}
\label{eq:guidedupdatewithst}
\eeq
Please refer to Fig.~\ref{fig:stvsxt} for a comparison of filtering $\restimate{t}$ versus $x_t$.

\subsection{Adjusting Guidance Strength}
\label{sec:adjustingguidance}
There are two ways to control the strength of guidance independent of our choice of filter $\guideprop$. The first is to adjust the guidance strength $\adjustmentweight$, and the second is to limit guidance to a particular range of iterations. We found little benefit to starting guidance late, but stopping guidance at some step $\tstop{}$ and letting the rest of the process proceed unguided is often useful. This is because edges and other coarse image structure, which tend to be the target of guidance, usually form early in the diffusion process, with later steps mostly adding finer detail to the image. The ideal guidance strength for a given image is subjective, but empirically, we found that fixing $\tstop$ around the last 20-30\% of steps and adjusting $\adjustmentweight$ provides a convenient way to explore different guidance strengths (see Fig.~\ref{fig:detail_ablation}).

\subsection{Filter Design}
\label{sec:filterdesign}
Our derivation so far has assumed that a filter $\guideprop{}$ can be chosen that separates coarse image structure from finer details. Fortunately, the design of such filters is well-explored in prior work. Of particular relevance is the bilateral filter ~\shortcite{bilateraltomasi}, which was used for style transfer before GANs ~\shortcite{BilateralStyleTransfer06}. More specifically, the joint bilateral filter~\cite{jointbilateral,fnoflash}, is a variant that can be used to selectively blur one signal based on the structure of another second guide signal. This is commonly used for guided upsampling (e.g., of depth maps based on corresponding high-resolution RGB data), and in our case, we will use it to derive a linear operation $\guideprop{}$ that captures the structure of our guide image. Normally, bilateral filters can be very expensive to evaluate. We address this by constructing what we call a \emph{joint bilateral tensor}.

\subsection{Joint Bilateral Tensor} 
\label{sec:fastbilateraltensor}
What typically makes the bilateral filter so expensive is a non-linearity related to the kernel's dependence on its own input. In contrast with a standard convolution filter, the bilateral kernel varies spatially and has a non-linear normalization term.
However, in our case, while the input to $\guideprop$ changes at every diffusion step, its dimensions and our guide image do not. This means that if we condition on our guide image, every output value becomes a fixed linear combination of input values. Leveraging this observation, we can unroll our filter into the product of our vectorized input image with a single square tensor. The width and height of this tensor are equal to the number of pixels in our input, which may be impractical at most standard image resolutions---this may be why, to our knowledge, we are the first to employ this strategy. However, most diffusion networks already operate on low spatial resolution (typically 64x64) latent images. At this resolution, the tensor fits easily into memory, letting us calculate a joint bilateral with arbitrary kernel size in a single near-instantaneous tensor operation. We calculate the tensor once at the start of diffusion, which costs approximately the same as running a single bilateral filter on one 64x64 image, adding very little overhead to our method.

The bilateral filter has two parameters: a spatial standard deviation $\sigma_{s}$, which controls the scale of blur, and a value standard deviation $\sigma_{v}$ which controls the preservation of edges. Fig.~\ref{fig:teaser} (left) shows how we can use these parameters to control guide structure. Increasing $\sigma_{s}$ allows for coarser-scale change, while decreasing $\sigma_{v}$ causes greater preservation of edge structure.

\subsection{Normalization for Color Distribution Shifts}
\begin{figure}
\begin{center}
    \includegraphics[width=1\linewidth]{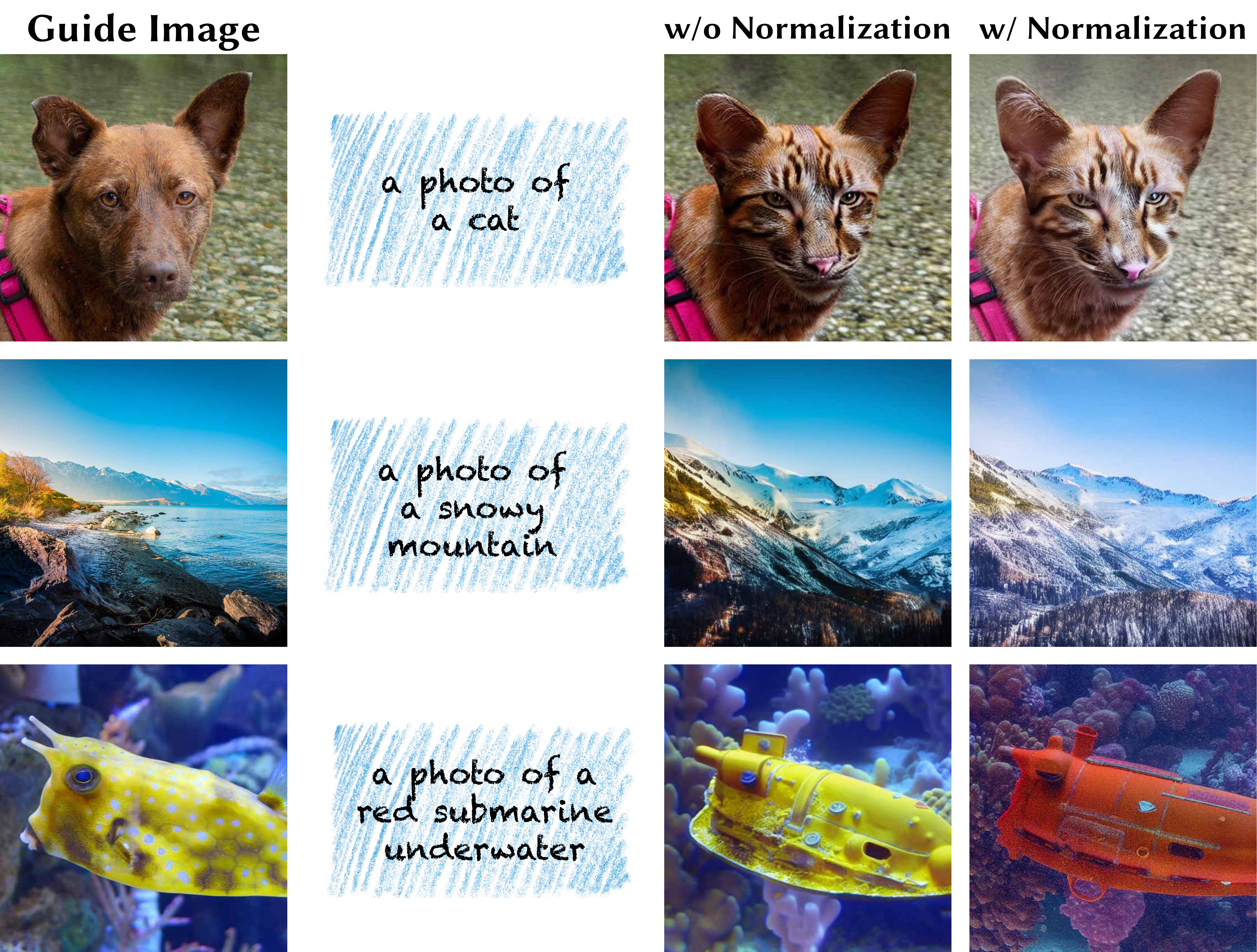}
\end{center}
    \caption{\textbf{Normalization for Color Distribution Shifts.} We show 3 typical scenarios for normalization: In the first row, FGD without normalization closely preserves the overall color of the guide image which is often desirable when the color from the guide image is a good match for the prompt such as translating between dogs and cats. In the second row, adding normalization improves the result by allowing for colors that better suit the prompt. In the third row, normalization is necessary to transform the color in order to satisfy the full prompt.}
\label{fig:norm_transition}
\end{figure}

\label{sec:methodnormalization}
We add one more improvement to optionally decouple the guidance of structure from the specific distribution of colors in a guide image. Up until now, we have not distinguished between the shape of a guide image and its color. Any low-pass guide filter will constrain the local color of output images just as much as their shape, which is undesirable when the distribution of guide colors is a poor match for a provided prompt. We can relax this constraint by re-normalizing our guide structure $\guidepropof{\guideim{}}$ at each step to match the means and standard deviations of whatever structure $\guidepropof{\rpreguideestimate{t}}$ it replaces. 
This lets the diffusion process shift the overall color distribution while respecting edge information from the guide image. Please refer to Algorithm 1 in our supplemental for precise equations.}

\subsection{Localized Edits}
\label{sec:localedit}
\begin{figure}[!h]
\begin{center}
    \includegraphics[width=\linewidth]{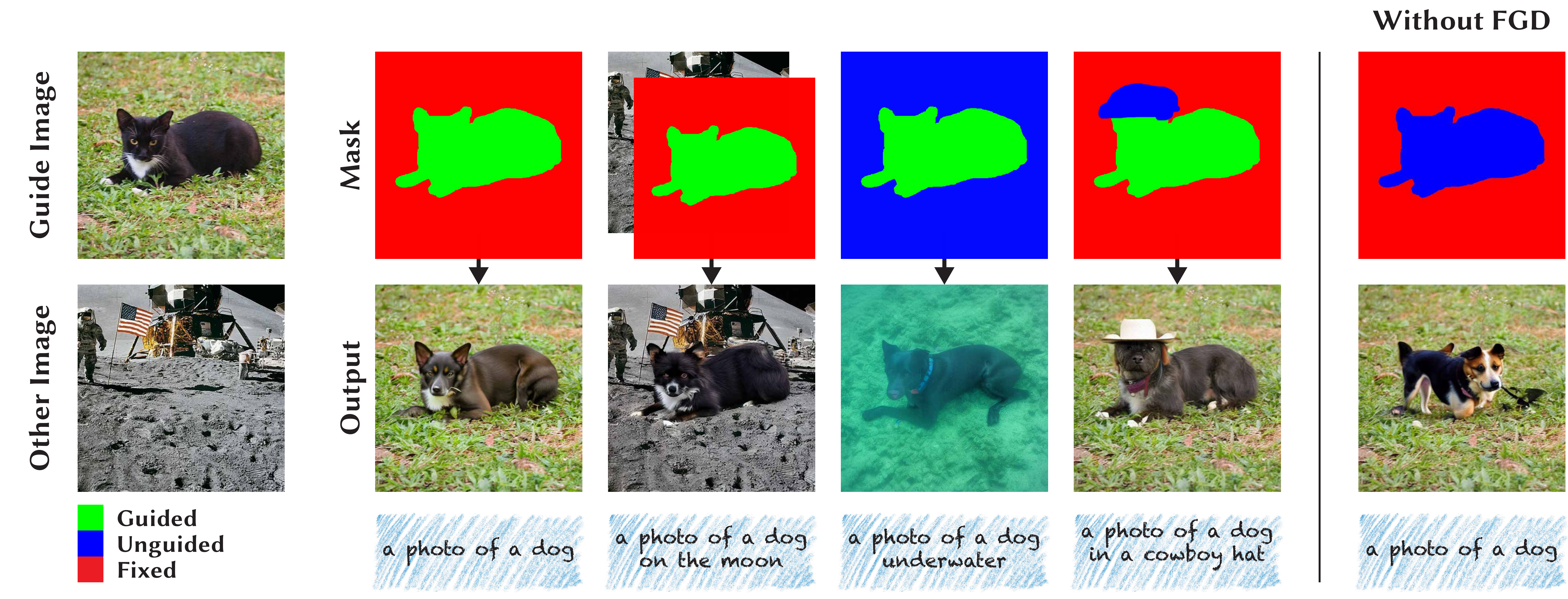}
\end{center}
    \caption{\textbf{Masking for Localized Edits.} 
    With masking, our method can make precise local edits specified by a user. For each mask, \emph{green} indicates where the image should be guided by FGD, \emph{blue} indicates where to give the model freedom to generate anything without guidance, and \emph{red} indicates where to strictly preserve the contents of the guide or other optional image. Notably, our masked guidance differs from in-painting, shown in the rightmost column, which does not adhere to the layout of the original cat.}
\label{fig:mask}
\end{figure}

The black box nature of FGD makes it easy to adapt for localized edits. The simplest way to do this is by masking the guidance vector applied at each step. In Fig.~\ref{fig:mask}, we show results based on replacing each $\rpreguideestimate{t}$ with a masked combination of three layers: one where FGD is applied, one where no guidance is applied, which can be used to give the network total freedom over part of an image, and  a third containing the unfiltered values of another image, which can be used to fix values in the final image. Different combinations of these layers can be combined to create a variety of different effects.

\subsection{Distinction from Other Black Box Methods}
\label{sec:distinctions_from_others}
At this point, we pause to note connections to other black box approaches. SDEdit can be expressed as a variation of FGD with $\guideprop{}=\mathbf{I}$ and $\tstop$ set to the $t_0$ parameter of SDEdit. Alternatively, if we choose $\guideprop{}$ to be a downsampling filter and use constant guidance strength $\adjustmentweight_{t}=1$, then Eq.~\ref{eq:guidedupdate} very closely resembles ILVR. The only difference is that ILVR injects extra noise at each step to keep $\rsample{t-1}$ in-distribution, which we can understand as compensating for the tendency of Eq.~\ref{eq:guidedupdate} to slightly denoise $\rsample{t-1}$. With this connection in mind, ILVR is closer to FGD but with several differences that account for improved performance. First, FGD allows the use of edge-preserving filters like our joint bilateral. Second, it offers more continuous control over guidance strength. Third, filtering denoised estimates $\restimate{t}$ instead of $\rsample{t}$ removes the need to re-inject extra noise at each step. And finally, our normalization strategy lets the generated color distribution adapt to a given prompt (see Fig.~\ref{fig:norm_transition} and Fig.~\ref{fig:fullpage_1}).
\begin{figure}
\begin{center}
    \includegraphics[width=\linewidth]{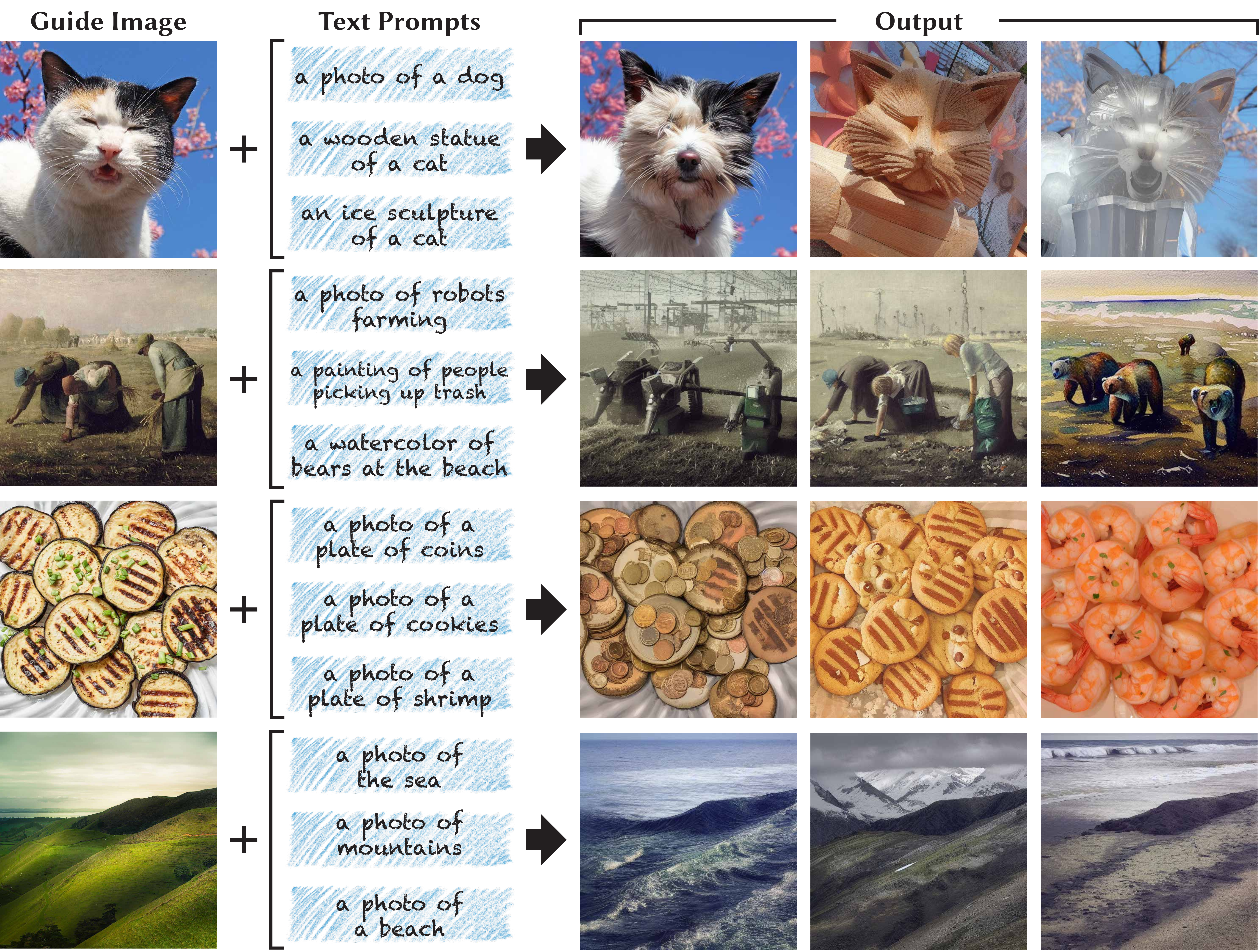}
\end{center}
    \caption{\textbf{FGD Results.} Our method consistently gives high quality translations across different materials, objects, and styles while respecting the structure from the guide image. For each row, the prompts from top to bottom correspond to the results from left to right. Please zoom in to view the pleasant details of our results.
    }
\label{fig:results_standalone}
\end{figure}

\section{Evaluation}
\label{sec:results}
\begin{figure*}
\centering
    \includegraphics[width=0.85\textwidth]{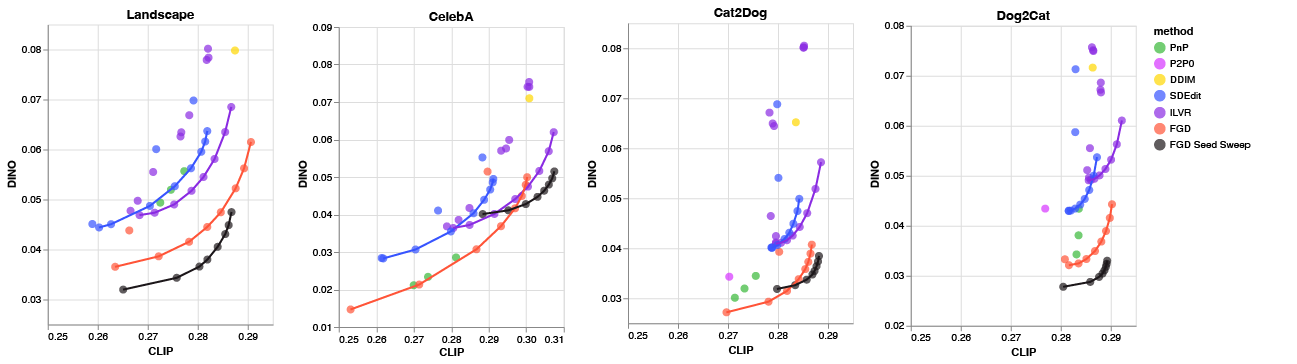}
    \caption{\textbf{Numerical Evaluation.} The goal is to balance a high CLIP similarity score with a small DINO difference. While there is no perfect metric for evaluating this balance, methods that perform well should be near the bottom right corner of the plot. For each method, a single point represents the average score of a single set of hyperparameter values, and the connected curve represents the changing L2 metric over the sweep of 7 hyperparameter sets described in Sec.~\ref{sec:results}. Our L2 weighting $\alpha$ ranges from 0.05 to 1.75, evenly spaced by 0.25. The black curves are sweeping over seeds using the optimal value set, and curves in other colors are sweeping over hyperparameter values. In both scenarios, our method achieves the best performance, using less time than the time one PnP run takes.
    To ensure fairness, we show multiple results for each compared method with different values of their hyperparameters. For PnP, we set the $t_{end}$ of their attention injection to be 20, 30, and 40. For ILVR, we set the $t_{end}$ to 20, 10, and 5, and the downsampling factor $N$ to 4, 8, 16. For SDEdit, we set the strength to 0.6, 0.75, and 0.85. To keep the plots clean, for single points without sweeping curves we visualize only the optimal hyperparameter set for FGD.
    }
\label{fig:numerical_all}
\end{figure*}

We compare against three black box methods: SDEdit~\shortcite{meng2021sdedit}, DDIM~\shortcite{song2020denoising},  and ILVR~\shortcite{choi2021ilvr}; as well as two SOTA white box methods: P2P0~\shortcite{parmar2023zero} and PnP~\shortcite{tumanyan2022plug}. All use SD v1--4~\shortcite{rombach2022high}.
As P2P0 requires additional text to estimate editing directions, we limit its evaluation to tasks for which the authors provided these (``cat'' to ``dog'' and vice versa).
We conduct extensive quantitative experiments on 3 datasets widely used for I2I translation tasks: CelebFaces Attributes (CelebA)~\cite{liu2015faceattributes}, Animal Faces-HQ (AFHQ)~\cite{choi2020stargan} and Landscape Pictures~\cite{afifi2021histogan}. Please see our supplemental for dataset details. In total, our dataset for quantitative evaluation consists of 1935 text-image pairs. For qualitative comparisons, we manually selected prompts and applied them to images found under Creative Commons licenses on the Internet.

\subsection{Computation Cost}
In Table~\ref{table:runtime}, we show the computation costs of FGD alongside other methods. In comparison to the most lightweight method SDEdit, our method adds only approximately 67 MB in memory, which is roughly $1/3$ of the memory required by PnP or P2P0. In terms of speed, our method adds only around 10 seconds, which is more than 10$\times$ faster than PnP and 2$\times$ faster than P2P0. Therefore, FGD can run at least 10 times within the time budget of one PnP run. 
\begin{table}[!h]
\caption{
\textbf{Computational Cost of Different Methods.}
For comparison, we ran each method on a NVIDIA A5000 GPU and averaged the results over a subset of our AFHQ dataset. The overhead time includes both loading the model as well as pre-computations that are specific to each method. For example, PnP spends a significant amount of time saving and loading attention features during this stage. In contrast, the only overhead time added by our method is for computing the joint bilateral tensor, which takes only 6.41 seconds (the remaining 6.51 seconds is for loading the model, acquiring the latent features of the guide image, etc.).
}
\begin{center}
\begin{tabular}{l r r r r} 
\hline
 Method & 
 \renewcommand{\arraystretch}{0.85}
 \begin{tabular}{@{}c@{}} Overhead \\ Time (\emph{s})\end{tabular} &
 \renewcommand{\arraystretch}{0.85}
 \begin{tabular}{@{}c@{}}Sampling \\ Time (\emph{s})\end{tabular} &
 \renewcommand{\arraystretch}{0.85}
 \begin{tabular}{@{}c@{}}Total \\ Time (\emph{s})\end{tabular} &
 \renewcommand{\arraystretch}{0.85}
 \begin{tabular}{@{}c@{}}Total \\ Memory ({MB})\end{tabular} \\
 \hline
 SDEdit & 4.44 & 5.61 & 10.05&5314.41\\ 
PnP & 241.25  & 9.14 & 250.39 & 14857.41\\
P2P0 & 18.19  & 36.00 & 54.19 & 18104.60\\
 FGD & 12.92 & 7.21&20.13 &5381.85\\ 
\end{tabular}
\label{table:runtime}
\end{center}
\end{table}

\subsection{Quantitative Results}
Following the practice of previous works, we evaluate on 2 competing objectives: (1) satisfying the text prompt, and (2) preserving the input guide structure. We use the same metrics as in PnP: CLIP score~\cite{radford2021learning}, and Structure Dist~\cite{tumanyan2022splicing} with a pre-trained vision transformer (DINO). The former measures the cosine distance between the feature of the edited results and the feature of the text prompt, so a higher value suggests better prompt adherence. As for DINO, a lower difference score means that the structure of the edited image is more similar to the input image.
Therefore, when plotted with CLIP similarity on the x-axis and Structure Dist on the y-axis, methods that land in the bottom right of the plot are estimated to perform better. 

As discussed in Sec.~\ref{sec:methodnormalization}, normalization is useful for translations that require significant color changes. Therefore, we turn it on for Landscapes which requires large color changes, and turn if off for AFHQ and CelebA (as skin-tone or fur color can be preserved). We fix $\adjustmentweight=1.6$ with normalization, and $\adjustmentweight=1.0$ without, and $t_{end}=10$.  

In Fig.~\ref{fig:numerical_all}, we use 7 sets of bilateral filter parameters (i.e. $\sigma_{s}$ and $\sigma_{v}$) for a single seed, and each point represents the average score for one hyperparameter setting. Instead of using up the PnP time budget entirely, we only run 7 times to stay well within the budget to allow for possible measurement errors. To choose the optimal result for each text-image pair, we calculate an overall score by 
\beq
l({\alpha})=\sqrt{\alpha*(CLIP-0.36)^2+(2-\alpha)*(DINO-0)^2}
\label{eq:l2_metrics}
\eeq
as 0.36 and 0 are the optimal magnitude CLIP and DINO limit to based on our observation on numerous results. To account for possible errors in these empirical limits, we weigh the two scores with a variety of $\alpha$ values before summing for the final L2 distance, i.e. creating different versions of the L2 metric, thus creating an optimal curve instead of a single point. One point on the optimal curve is an average over the optimal result of each text-image pair, selected by the L2 distance with one certain CLIP--DINO weighing ratio. We call running over multiple settings and picking the optimal results as ``sweeping''. Another way to exploit diversity is to sweep over different seeds with a single optimal set of hyperparameters, so we draw another curve by evaluating on 7 seeds. We also present the optimal curves for other fast black box methods. Assuming some smoothness over the change of CLIP--DINO ratio, our curve gives a lower DINO when having the same CLIP score, or a higher CLIP when having the same DINO score, suggesting strict superiority. Note that for Dog2Cat and Cat2Dog, we beat P2P0 with only a single hyperparameter set, which saves us from doing the optimal sweep with its time budget.

\subsection{Qualitative Results}

\label{sec:qualitativeresults}
\begin{figure}
\begin{center}
    \includegraphics[width=1\linewidth]{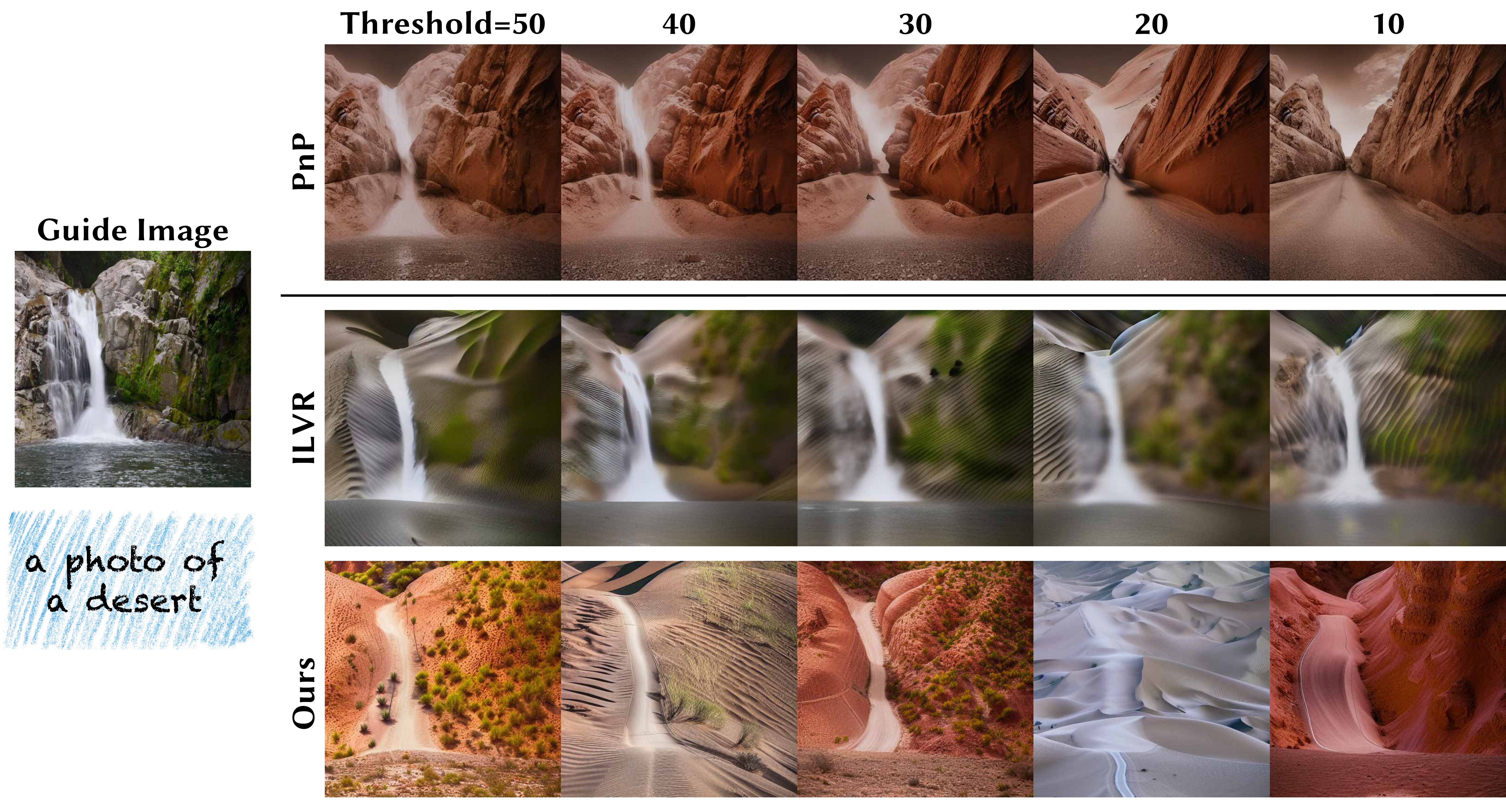}
\end{center}
    \caption{\textbf{Diversity Comparison.} Although one hyperparameter of PnP, the feature injection threshold, controls translation similarity, it lacks versatility due to being bounded with deterministic DDIM. In comparison, non-deterministic FGD produces diverse high-quality results across different seeds. We also ran ILVR across different seeds but observe their method has smoothing artifacts and cannot deviate from the color of the guide image which limits variety and prompt adherence. To further disentangle measurements of quality and diversity, we measure FID$\downarrow$~\cite{heusel2017gans} on all our datasets over several seeds with optimal hyperparameters and found ILVR averages 3.19 times higher than FGD.
    }
\label{fig:diversity}
\end{figure}

As different text-image pairs require different structural changes, it is not surprising that sweeping over different hyperparameter values is helpful. Yet sweeping has additional benefits for non-deterministic methods as different seeds create an additional mode of generation.
For qualitative results shown in the paper, we generate our results using a fixed pool of two different sets of hyperparameters over three seeds which is completed in around half the time of a single PnP run. For other white box methods, we generate their results with the optimal values suggested in their papers. Since these are bound to DDIM, seeding has a significantly smaller impact as shown in Fig.~\ref{fig:diversity}.

We show qualitative results of FGD on a fixed guide image with different texts in Fig.~\ref{fig:results_standalone}, and more comparisons with other methods in Fig.~\ref{fig:fullpage_1}. In Fig.~\ref{fig:diversity}, we compare the generation diversity with a top-performing white box method, PnP, and black box method ILVR. Even though the latter uses non-deterministic sampling, its diversity is limited by the entanglement of structure and color.

\subsection{Ablation}
\begin{figure}
\begin{center}
    \includegraphics[width=0.8\linewidth]{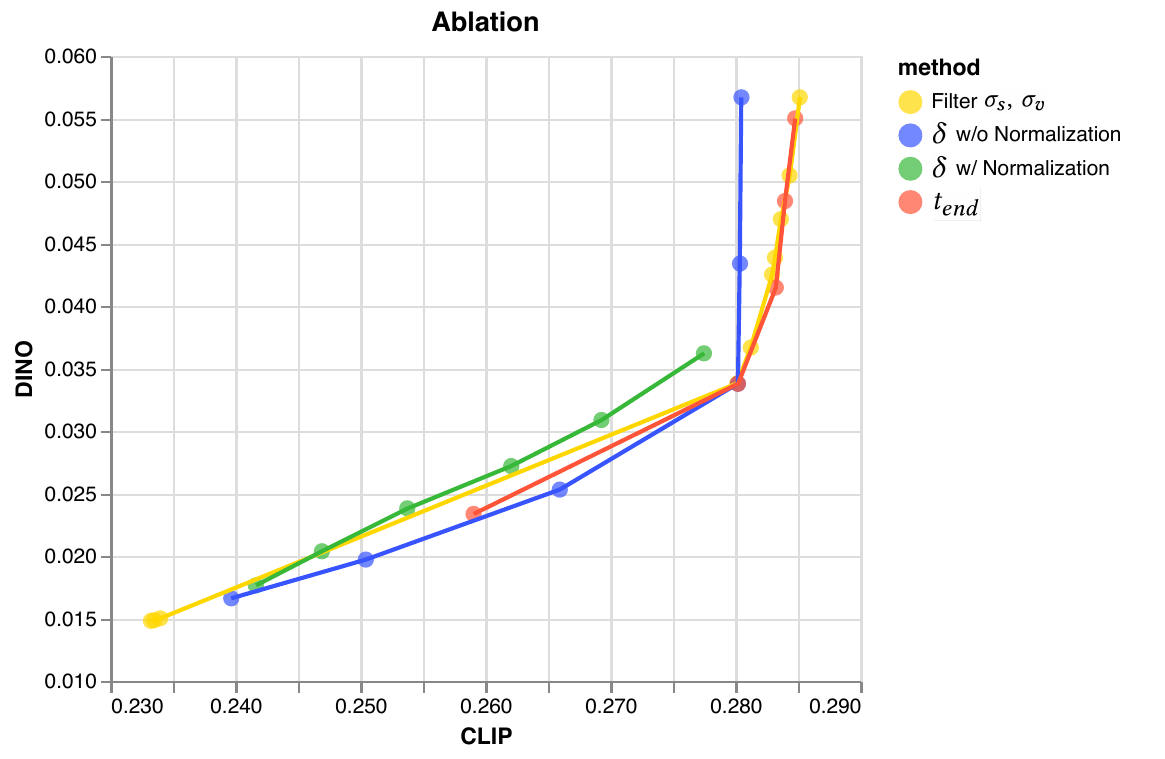}
\end{center}
    \caption{\textbf{Quantitative Ablations.} We ablate on the hyperparameters used by FGD: $\sigma_{s}$, $\sigma_{v}$, $\adjustmentweight$, $t_{end}$, and whether the color normalization is used. We run $\adjustmentweight=0.6, 0.8, 1.0, 1.2, 1.4, 1.6$ without normalization, and $t_{end}=25, 20, 15, 10, 5$. Additionally, we sweep over $\adjustmentweight=1.0, 1.2, 1.4, 1.6, 1.8, 2.0, 2.2$ with normalization: the structure is still well-controlled, but overall the results are better without it, which along with Fig.~\ref{fig:norm_transition} support our understanding of the usage of normalization, and shows that it is not necessary for the Dog2Cat dataset.}
\label{fig:ablations}
\end{figure}

\label{sec:ablations}
We ablate all our hyperparameters ($t_{end}$, $\adjustmentweight$, $\sigma_{s}$, $\sigma_{v}$, and normalization) on a 300 subset of the Dog2Cat AFHQ dataset. We first show the influence of $\adjustmentweight$: increasing it increases the structure coherence, and when if gets too strong, the CLIP score would be sacrificed as expected. Then we fix $\adjustmentweight$ to be the optimal 1 and sweep over $t_{end}$: similarly, the more steps we apply our filter, the more the output adheres to the guide structure. In the end, we sweep over various $\sigma_{s}$ and $\sigma_{v}$ combinations to show the optimal set of hyperparameters for this dataset is $t_{end}=10$, $\adjustmentweight=1$, $\sigma_{s}=3$, $\sigma_{v}=0.3$.

\section{Limitations}
\label{sec:limitations}
While FGD results are competitive or superior to current single-shot attention-based approaches, black box methods have some fundamental limits in the long-term. Access to internal features and network gradients can provide useful information, especially when the distribution of images with given guide structure is very different from the distribution meeting a given text prompt. On the other hand, FGD can be more effective when one wants to express control explicitly in terms of appearance.

Although FGD is fast compared with other current methods, gradual guidance depends on the iterative nature of diffusion. If networks or samplers seek to speed up the diffusion in the future by reducing the number of iterations, this could interfere with FGD's guidance past some point. Our supplemental material shows results of FGD with samplers other than the DDPM, indicating some robustness to such changes, yet this issue could still become a limit for potential sampler or network designs in the future.
\section{Conclusion}
This work presents a simple and effective black box approach for adding high-quality image-based guidance to image diffusion models. Our FGD framework is flexible, and while our results focus on guide filters from the joint bilateral family, we believe exploring other filters will be an interesting avenue for future work. We also believe that the connections made with older \ITIT{} and image processing methods offer useful insight into the behavior of diffusion. Research in \ITIT{} is rapidly evolving, but the simplicity, speed, and generality of FGD make it an especially good baseline or plugin for adding image-based guidance to other methods.

\begin{acks}
This work was supported in part by a generous gift from Meta. We also thank Nhan Tran for helping us make the submission supplemental page.
\end{acks}

\newpage
\bibliographystyle{ACM-Reference-Format}
\bibliography{main}

\appendix

\begin{figure*}
\centering
    \includegraphics[width=\textwidth]{figures/full_page_wide.pdf}
    \caption{\textbf{More Results.} We recommend zooming in to see the different quality of details across methods. For example, observe our method is able to apply the color transformation in the second to last row while also preserving the smaller fish while other methods struggle with one or both of these tasks. }
\label{fig:fullpage_1}
\end{figure*}

\end{document}